\documentclass[runningheads]{llncs}
\usepackage{graphicx}
\usepackage{tikz}
\usepackage{comment}
\usepackage{amsmath,amssymb} % define this before the line numbering.
\usepackage{color}
\usepackage{hyperref}
\usepackage{orcidlink}
\usepackage[accsupp]{axessibility}  % Improves PDF readability for those with disabilities.
\usepackage{longtable}
\usepackage{makecell}

\begin{document}
% \linenumbers
\pagestyle{headings}
\mainmatter
\def\ECCVSubNumber{149}  % Insert your submission number here

\title{Spatial Temporal Graph Attention Network for Skeleton-Based Action Recognition} % Replace with your title

\begin{comment}
\titlerunning{ECCV-22 submission ID \ECCVSubNumber} 
\authorrunning{ECCV-22 submission ID \ECCVSubNumber} 
\author{Anonymous ECCV submission}
\institute{Paper ID \ECCVSubNumber}
\end{comment}
%******************

%\begin{comment}
\titlerunning{STGAT for Skeleton-Based Action Recognition}
\author{Lianyu Hu\inst{1}\orcidlink{0000-0003-2470-8110} \and
Shenglan Liu\inst{2}\orcidlink{0000-0003-3823-4200} \and
Wei Feng\inst{1}\orcidlink{0000-0003-3809-1086}}
\authorrunning{Lianyu et al.}
\institute{Tianjin University \and Dalian University of Technology
%\email{hly2021@tju.edu.cn;liusl@dlut.edu.cn;wfeng@ieee.org}
}
\vspace{-20px}
%\end{comment}
%******************
\maketitle

\begin{abstract}
It's common for current methods in skeleton-based action recognition to mainly consider capturing long-term temporal dependencies as skeleton sequences are typically long (\textgreater 128 frames), which forms a challenging problem for previous approaches. In such conditions, short-term dependencies are few formally considered, which are critical for classifying similar actions. Most current approaches are consisted of interleaving spatial-only modules and temporal-only modules, where direct information flow among joints in adjacent frames are hindered, thus inferior to capture short-term motion and distinguish similar action pairs. To handle this limitation, we propose a general framework, coined as STGAT, to model cross-spacetime information flow. It equips the spatial-only modules with spatial-temporal modeling for regional perception. While STGAT is theoretically effective for spatial-temporal modeling, we propose three simple modules to reduce local spatial-temporal feature redundancy and further release the potential of STGAT, which (1) narrow the scope of self-attention mechanism, (2) dynamically weight joints along temporal dimension, and (3) separate subtle motion from static features, respectively. As a robust feature extractor, STGAT generalizes better upon classifying similar actions than previous methods, witnessed by both qualitative and quantitative results. STGAT achieves state-of-the-art performance on three large-scale datasets: NTU RGB+D 60, NTU RGB+D 120, and Kinetics Skeleton 400. Code is released.
%\keywords{Spatial temporal modeling, Graph attention network, Recognition on similar actions }
\end{abstract}
  
\section{Introduction}
\label{sec:sec1}
As a special issue of action recognition, skeleton-based action recognition has attracted much attention recently due to the fast development of pose estimation algorithms~\cite{cao2017realtime,fang2017rmpe} and wearable motion capture equipment. It involves predicting an action label from a sequence of 2D or 3D skeleton representations. Compared to action recognition in videos~\cite{wang2018non,wang2016temporal,carreira2017quo}, skeleton data is more compact in 2D~\cite{kay2017kinetics,yan2018spatial} or 3D~\cite{liu2019ntu,shahroudy2016ntu} format and suffers less from environmental factors (e.g. camera motion, light transformations and viewpoint changing), which makes it a more robust representation of human action sequences.
  
\begin{figure}[t]
\centering
\includegraphics[width=0.8\columnwidth]{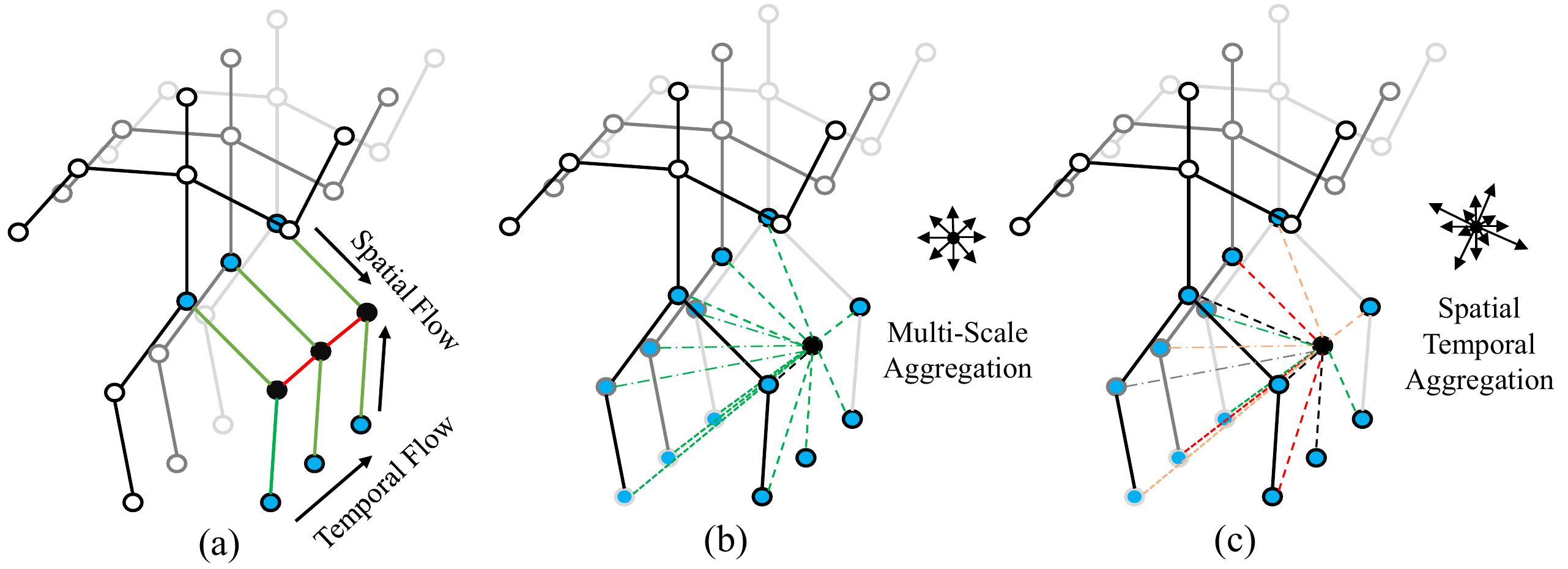} 
\caption{(a) Factorized spatial-only and temporal-only modeling over input sequences. (b) Disentangled multi-scale feature aggregation in local spatial-temporal neighborhoods. Edges of the same color indicate equal weights in the predefined spatial-temporal graph. (c) Spatial-temporal feature aggregation with adaptively computed edges. Edges with different colors represent different degrees of connection strength.} %Best viewed with color.}
\label{fig1}
\vspace{-15px}
\end{figure}

It's usual for current methods to mainly tackle long-term temporal dependencies as skeleton sequences are rather long (\textgreater 128 frames), which forms a serious obstacle for previous methods to model human dynamics through a whole video. However, in such conditions, critical short-term temporal dependencies are few formally considered. In contrast with long-term dependencies which recognize action sequences from an \textit{overall} perspective, short-term dependencies distinguish similar sequences on \textit{small} temporal scales with important details. For example, 'Eating' and 'Tooth brushing' have quite similar movements in both upper and lower body where hands are raised next to mouth for a long while, holding food or brushing teeth, which is quite similar in long sequences. To classify them relies on whether there exists subtle movement of the hand to brush horizontally, which is short-term and rather critical. Few methods~\cite{liu2020disentangling,zhang2020spatial} have realized the importance of short-term motion and proposed dilated temporal convolutions to capture short-term temporal movements. We propose a \textit{general} framework to perform spatial-temporal modeling, achieving superior performance than them under similar computational budgets as shown in Sec.~\ref{sec:sec4.3}.
  
Most existing methods employ interleaving spatial-only and temporal-only modules to extract spatial features and capture long-term temporal dependencies, respectively. A typical approach is to first use a graph module to extract spatial representations at each time step, and then deploy recurrent or temporal convolution layers to build temporal relationships (Fig.\ref{fig1}(a))~\cite{li2019spatio,si2019attention,li2019actional,shi2019two,yan2018spatial}. Such design are efficient and effective for long-term modeling, but hinders direct information flow in local spatial-temporal neighborhoods. For example, messages have to pass through a spatial module and a temporal module to reach nodes in adjacent frames, which are \textit{inevitably} weakened in multiple transitions. Especially, large temporal kernels adopted after spatial modules get more information mixed and may cause subtle motion impaired, where short-term temporal dependencies may be ineffectively captured by such factorized modeling.
  
In this work, we propose STGAT with spatial-temporal modeling to specially capture short-term dependencies. We retain the temporal-only modules for long-term modeling and give spatial-only modules more freedom to perform local spatial-temporal modeling. It builds local spatial-temporal graphs by connecting nodes in local spatial-temporal neighborhoods and dynamically constructing their relationships (Fig.\ref{fig1}(c)). Thus messages can directly reach other joints in local neighborhoods without multiple transmission costs. The connection strength of edges is adaptively determined to better fit local movement patterns. For further releasing the potential of STGAT, we propose several light-weight modules to reduce local feature redundancy. These modules make effect in narrowing the scope of self-attention mechanism, dynamically weighting joints along temporal dimension and separating subtle motion from static features. Equipped with the ability in spatial-temporal modeling, STGAT exhibits great progress on classifying similar action pairs, witnessed by both qualitative and quantitative results. As a result, STGAT achieves state-of-the-art performance on three large-scale datasets: NTU RGB+D 60~\cite{shahroudy2016ntu}, NTU RGB+D 120~\cite{liu2019ntu}, and Kinetics Skeleton 400~\cite{kay2017kinetics}. Code is released\href{https://github.com/hulianyuyy/STGAT}{https://github.com/hulianyuyy/STGAT}.

\section{Related work}
\subsection{Skeleton-Based Action Recognition}
Earlier methods always make use of hand-crafted features~\cite{vemulapalli2014human,wang2012mining} to depict human dynamic representations. However, they fail to model dependencies across spacetime and lack enough ability in extracting high-dimensional features. Later, convolutional neural network(CNNs) and recurrent neural networks(RNNs) become mainstream by modeling human dynamics as a pseudo-image~\cite{ke2017new,li2017skeleton,li2018co} or a series of coordinates along time~\cite{du2015hierarchical,liu2016spatio,li2018independently,zhang2017view,si2019attention}. Nevertheless, they overlook the internal relationships between joints which are better captured by graph networks due to their natural advantage over handling non-euclidean data. Graph networks in skeleton-based action recognition fall into two streams: spatial perspective and spectral perspective. ST-GCN~\cite{yan2018spatial} belonging to the spatial perspective firstly models human structure as a spatial graph to construct node relationships. Later methods mainly make efforts in enlarging the receptive field~\cite{gao2019optimized,li2019actional,shi2019two,chen2021multi}, combining another stream~\cite{shi2019two,li2019actional,liu2020disentangling,shi2019skeleton}, dividing graphs into structural ones~\cite{li2019actional,thakkar2018part}, design network structure~\cite{huang2020spatio,peng2020learning,song2020stronger,chengdecoupling,chen2021multi}, combining adaptive learning~\cite{shi2019two,ye2020dynamic,shi2019skeleton,zhang2020semantics,hu2020dual,shi2020decoupled,bai2021gcst}.
  
\subsection{Self-Attention Mechanism with Skeleton Data}
One of the main benefits of self-attention mechanism is to adaptively adjust relationships with neighbors according to their responses. A self-attention module flexibly computes the response of a position with others' in an embedding space and then aggregates their weighted features. Self-attention mechanism has proven its power in many fields ~\cite{cheng2016long,lin2017structured,wang2018non,velivckovic2017graph} and is recently introduced to model relationships~\cite{ye2020dynamic,shi2019skeleton,shi2019two,liu2020disentangling,zhang2020semantics,shi2020decoupled} in skeleton-based action recognition. 2s-AGCN adds an adaptively computed graph with the physically predefined graph to flexibly model joint relationships, while MS-G3D~\cite{liu2020disentangling}, CA-GCN~\cite{zhang2020context} and dynamic-GCN~\cite{ye2020dynamic} inherit it. DSTA-Net~\cite{shi2020decoupled} and SGN~\cite{zhang2020semantics} proposes to alternatively employ fully attention-based spatial modeling and temporal modeling for capturing movement patterns. In contrast with previous approaches, our method \textit{doesn't} rely on a certain self-attention strategy, but proposes a spatial-temporal modeling framework for capturing long-term and especially short-term dependencies.
  
\section{Learning STGAT}
\subsection{Preliminaries}
\subsubsection{Notations.}
Formally, a graph is always represented as \textit{G}=(\textit{V}, \textit{E}), where \textit{V} = \textit\{$v_{1}$,...,$v_{N}$\} is a set of \textit{N} graph nodes representing joints and \textit{E} is a series of graph edges representing bones between joints. An adjacent matrix \textit{A} with size \textit{N}$\times$\textit{N} is adopted to depict bones where $A_{i,j}$ represents the connection strength between node \textit{i} and \textit{j}, which equals zero if they have no connections. As for input, action sequences have a node features set represented as a feature tensor $X^{C\times T\times N}$ where each node $v_{n}\in$ \textit{N} has a \textit{C} dimensional feature vector over total \textit{T} frames.
\subsubsection{Graph Convolutional Networks.}
Given the input tensor \textit{X} and graph structure \textit{A}, layer-wise graph convolution at each time step can be implemented as:
\begin{equation}
\label{e1}
X^{out}=\sigma(WX^{in}\Lambda^{-\frac{1}{2}}(A+I)\Lambda^{-\frac{1}{2}})
\end{equation}    
where $X^{out}$ and $X^{in}$ separately corresponds to the output and input features and \textit{W} is a trainable weight matrix. \iffalse Only 1-hop neighbors of \textit{A} have the same normalized non-zero value.\fi $\Lambda$ is the diagonal degree matrix of \textit{A} to employ normalization and \textit{I} adds self-loops to keep identity features. At last, $\sigma (\cdot)$ acts as an activation function for output. 
  
\subsection{Spatial-Temporal Graph Attention Network}
%\subsubsection{Current Problem of Handling Short-Term Dependencies}
Following the statement in Sec.~\ref{sec:sec1}, short-term dependencies are critical cues for classifying similar action pairs. We propose a general framework to capture short-term dependencies by performing spatial-temporal modeling, detailed as follows.
  
\subsubsection{Building Spatial-Temporal Graphs.}
To better capture short-term dependencies, in this paper we give spatial-only modules more freedom for short-term aggregation, while maintaining the role of temporal-only modules for long-term modeling. We extend spatial-only modules to build a local robust spatial-temporal feature aggregator which directly exchange messages between nodes in a local spatial-temporal neighborhood. We start by building a spatial-temporal graph consisting of nodes in a local cross-spacetime neighborhood, which have links with not only other nodes in their original \textit{spatial} graph but also their neighbors in \textit{neighboring} timestamps. In this way, each node can \textit{directly} aggregate messages from other local spatial-temporal neighbors. 
  
Let us consider a sliding window of size $\tau$ and dilation \textit{d} at each time step over input sequences, which generates a local action sequence at each time step \textit{t} represented as $X^{t}_{\tau}=\{x_{t-\tau/2:t+\tau/2}\in R^{C\times \tau N}|t\in Z, 0\le t < T\}$. Here, $\tau$ controls the local sequence length along time while \textit{d} represents the dilation rate by picking a frame every \textit{d} frames. Dilated window allows us to control the interval between sampled frames to fit local movement patterns. Then, for each frame $x_t$, a unique regionally spatial-temporal adjacent matrix $\widehat{A}^{t}_{\tau}$ is organized by enumerating all possible neighbors in local spatial-temporal neighborhood $X^{t}_{\tau}$ as:
\begin{equation}
\label{e2}
\widehat{A}^{t}_{\tau}=[\widehat{A}^{t}_{(1)},\cdot,\widehat{A}^{t}_{(\tau)}] \in R^{\tau N\times N}
\end{equation}
where $\widehat{A}^{t}_{(\tau)}$ represents a cross-time graph with size $N\times N$ by establishing relationships between nodes at current timestamp \textit{t} with their neighbors at $\tau$ neighboring frames in a local spatial-temporal neighborhood. If we take one node of current timestamp \textit{t} as an example (i.e. one column of $\widehat{A}^{t}_{\tau}$), it's connected to $\tau\times N$ neighbors consisting of  $\tau$ frames with \textit{N} nodes at each timestamp. By incorporating $\widehat{A}^{t}_{\tau}$ into graph operations, the output of STGAT for each frame $x_t$ can be obtained as:
\begin{equation}
\label{e3}
[X^{out}]^t=\sigma(W[X_{\tau}^{in}]^t\widehat{A^{t}_{\tau}}).
\end{equation}
\subsubsection{Adaptive Computation for Spatial-Temporal Graph.}
For well distinguishing similar action pairs, more attention should be paid to aggregate messages from important joints and less to other joints. We incorporate self-attention mechanism to \textit{adaptively} compute edge weights in local spatial-temporal graphs to achieve this goal. 

Following the widely used self-attention formulation, a generic self-attention operation in graph networks at each time step can be defined as:
\begin{equation}
\label{e4}
\widehat{A}_{i}=\frac{1}{C(X_{i}^{in})}\sum_{\forall j}{f(X_{i}^{in},X_{j}^{in})}g(X_{j}^{in})
\end{equation}
where \textit{i} is the index of an output position, \textit{j} is the index that enumerates all possible neighbors and $\widehat{A}_{i}$ is the computed edge weights of node \textit{$v_i$} with others. Here, the pairwise function \textit{f}$(\cdot)$ computes the affinity of features between node \textit{$v_i$} and \textit{$v_j$}. \textit{g}$(\cdot)$ offers an embedding feature for node \textit{$v_j$}. $C(\cdot)$ adds normalization for the result. Given the adjacent matrix $\widehat{A}$, a weighted output $X^{out}$ can be computed as:
\begin{equation}
\label{e5}
X^{out}=\sigma(WX^{in}\widehat{A}).
\end{equation}  

A more stable and flexible choice is employing multi-head attention modules to learn various kinds of edge weights. Specially, \textit{S} independent self-attention modules are employed to learn different graph structures, as:
\begin{equation}
\label{e6}
X^{out}=\sigma(\frac{1}{S}\sum_{s=0}^{S}W_sX^{in}\widehat{A}_{s}).
\end{equation}
Here $\widehat{A}_{s}$ and $W_s$ are the computed adjacent matrix and weight matrix separately of $s_{th}$ head. 
  
By incorporating our spatial-temporal graph $\widehat{A}^{t}_{\tau}$ and our sampled local action sequence $X^{t}_{\tau}$ in Eq.~\ref{e5}, we arrive at the final form to compute edge weights and aggregate messages from neighbors in spatial-temporal graphs:
\begin{equation}
\label{e7}
X^{out}=\sigma(\frac{1}{S}\sum_{s=0}^{S}W_s[X^{in}]^{t}_{\tau}\widehat{A}^{t}_{\tau,s})
\end{equation}
where $\widehat{A}^{t}_{\tau,s}$ represents a \textit{unique} computed spatial-temporal graph at timestamp \textit{t} with temporal length $\tau$ of the $s_{th}$ head. $X^{out}$ is the accumulated output of all heads with different spatial-temporal graphs. 
  
Based on above design, our STGAT is a \textit{general} framework which doesn't rely on a certain self-attention strategy and can be built with self-attention operators or any attention-based approaches by inflating their spatial-only modules to construct cross-spacetime edges for short-term modeling. 
 
\subsubsection{Discussion.}
We give some in-depth analysis for STGAT as follows: (1) Compared with MS-G3D~\cite{liu2020disentangling} which builds a spatial-temporal graph with size $\tau N \times \tau N$ for cross-spacetime modeling, ours owns size $\tau N \times N$. Thus, MS-G3D introduces more redundant messages from neighbors and cost \textit{N} times of computations than ours. More importantly, MS-G3D assigns \textit{same} edge weights for joints in a local spatial-temporal neighborhood (fig.\ref{fig1}(b)), which fails to distinguish beneficial messages and allows too much information to flood into current node without filter. In contrast, we dynamically construct joint relationships and thus is able to aggregate representative features. In this condition, MS-G3D deploys two pathways and proposes multiscale temporal convolutions to capture various short-term motion but still performs worse than ours. 2. STGAT is analogous to 3DConv~\cite{tran2015learning,carreira2017quo} in modeling spatial-temporal dependencies by a weighted aggregation of regional cross-spacetime messages. While the parameters of 3D convolution kernels are structurally learned and fixed for samples, STGAT adaptively computes the edge weights for each sample and is less prone to overfit than 3DConv during training. 
  
\subsection{Reducing local spatial-temporal redundancy}
While STGAT is theoretically effective for spatial-temporal aggregation, we empirically found direct aggregation of messages from spatial-temporal neighbors brings redundant \textit{static} information, resulted from similar contents in adjacent frames. 
We propose several simple modules to reduce local spatial-temporal redundancy and further release the potential of STGAT, resulting in a robust feature extractor.
\subsubsection{Separate Learning.}
Compared to the original \textit{N}$\times$\textit{N} spatial graph, STGAT builds a $\tau$\textit{N}$\times$\textit{N} spatial-temporal graph to model local dependencies across spacetime. Obviously, the size of adjacent matrix increases $\tau$ times. That makes it harder for self-attention operators to adaptively construct joint relationships among neighbors with \textit{similar} contexts.  As we found in practice, as node number grows, attention-based modules tend to assign uniform weights around 1/($\tau\times N$) for each joint, which thus gradually loses the ability to adaptively construct relationships. We propose Separate Learning Strategy to limit node numbers and keep beneficial information \textit{lossless} to relieve this problem. Inspired by group-wise convolution~\cite{krizhevsky2012imagenet}, we propose to divide joints in a local spatial-temporal region into several small groups, where STGAT only weights edges within groups. Thus, the scope of self-attention operators is narrowed and the computation costs are proportionally lowered down as well. Features from different groups are concatenated as outputs which won't bring information loss. Practically, while there exist lots of methods to group joints, we choose to divide them according to their distances with target joint. 
  
\subsubsection{Dynamic Temporal Weighting.}
As defined in Eq.~\ref{e2} and ~\ref{e3}, STGAT builds spatial-temporal graphs to model cross-spacetime information. We notice that some keyframes play more \textit{important} roles than others where considering frames equally may hurt modeling short-term motion. We propose two methods to dynamically weight frames regionally as follows and compare them in the experiments:\\
(1) We add a learnable parameter $T_{DTW}$ with size \textit{S}$\times\tau$ to weight frames for \textit{S} heads with temporal duration $\tau$, which is updated through network propagation. Note that $T_{DTW}$ \textit{structurally} considers the weights of frames in each head. It's fixed for each sample during inference.\\
(2) Inspired by SENet~\cite{hu2018squeeze}, we design an adaptive temporal weighting module to weight frames by considering the short-term motion in a local spatial-temporal region. We first perform 1-hop graph convolutions for current frame $X^{t}$ to aggregate local spatial features. Then, a $1\times 1$ convolution kernel is utilized to enhance features of current frame $X^{t}$. We subtract features of regional neighbors from those of current frame to obtain motion features, which reflect the movement degree of neighboring frames against current one. Finally, after pooling along channel and temporal dimension, they are passed through a sigmoid function to weight frames in local spatial-temporal region. These weights are computed \textit{adaptively} in accordance with input for regional weighting. An Illustration can be found in the supplementary material.
\begin{figure}[t]
\centering
\includegraphics[width=0.65\columnwidth]{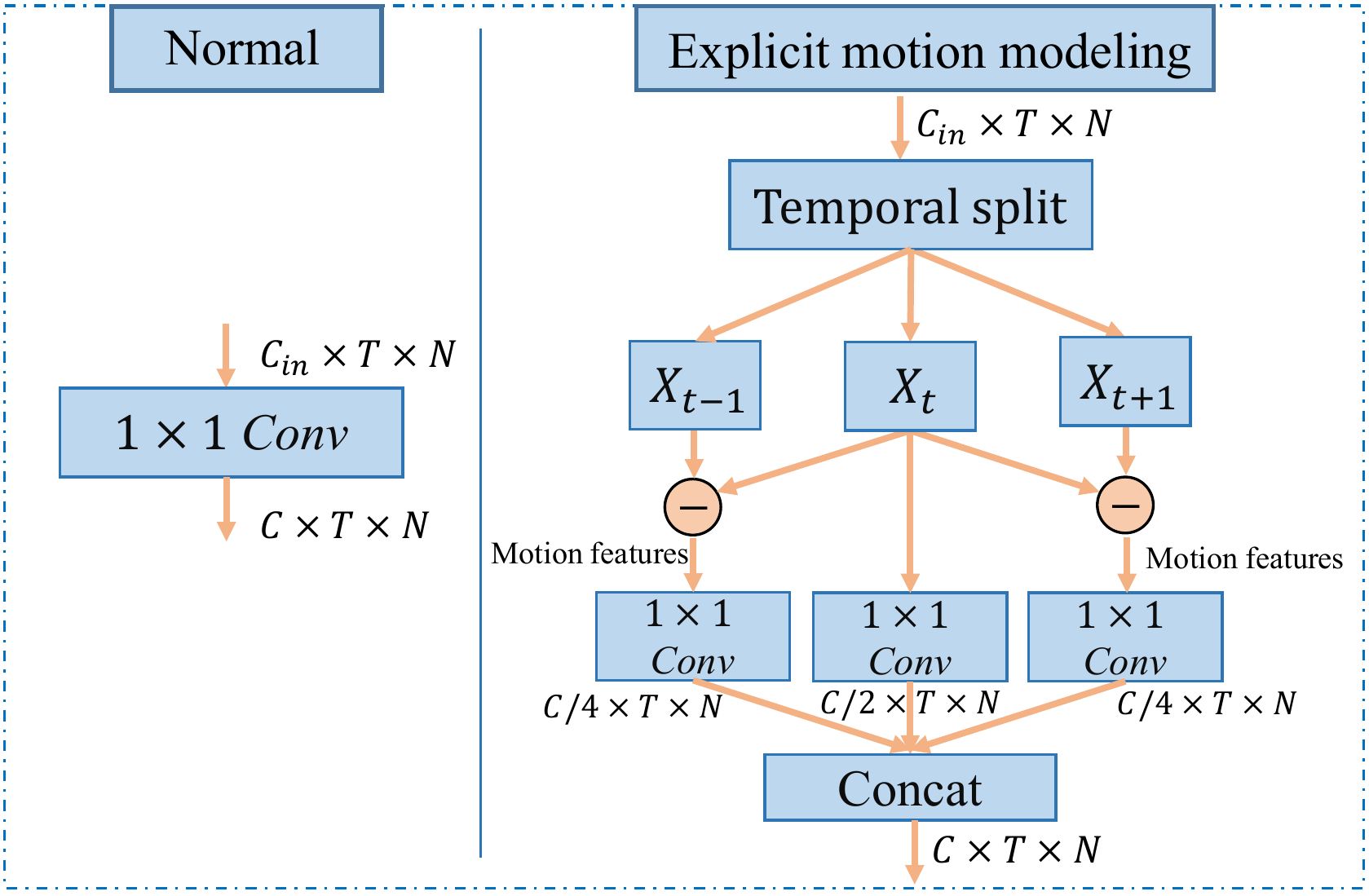} 
\caption{ Comparison of our proposed Explicit Motion Modeling strategy with normal $1\times 1$ implementation. We try to separate short-term motion in local spatial-temporal region from static features for better short-term modeling.}
\label{fig4}
\end{figure}
\subsubsection{Explicit Motion Modeling.}
Considering neighboring frames always contain similar contents, we infer only one frame can provide enough static information and short-term motion benefits recognition more, for reducing local feature redundancy. We propose Explicit Motion Modeling strategy to separate short-term motion in local spatial-temporal neighborhood from static features. As illustrated in fig.\ref{fig4}, compared to original $1\times 1$ transformations, we first split the features along temporal dimension. And then, we subtract features of adjacent frames from those of current frame to obtain motion features. In this condition, only \textit{static} features of \textit{current} timestamp is reserved and short-term motion is obtained. With appropriate padding, these motion features pass a $1\times 1$ convolution kernel for enhancement and are finally concatenated with current frame's features as output. Compared with normal $1\times 1$ convolution, this design introduces no additional parameters but enhances the ability for distinguishing short-term motion.
 
\vspace{-10px}
\subsubsection{Discussion.}
These three modules are lightweight and can be painlessly incorporated into STGAT, with \textit{quite little} costs. Only Dynamic Temporal Weighting consumes \textit{S}$\times\tau$ parameters, which are negligible in deep-learning-based networks. They make powerful effect for reducing local redundancy, resulting in superior performance of STGAT, as witnessed in sec.~\ref{sec:sec4.3}.
\begin{figure}[t]
\centering
\includegraphics[width=0.8\columnwidth]{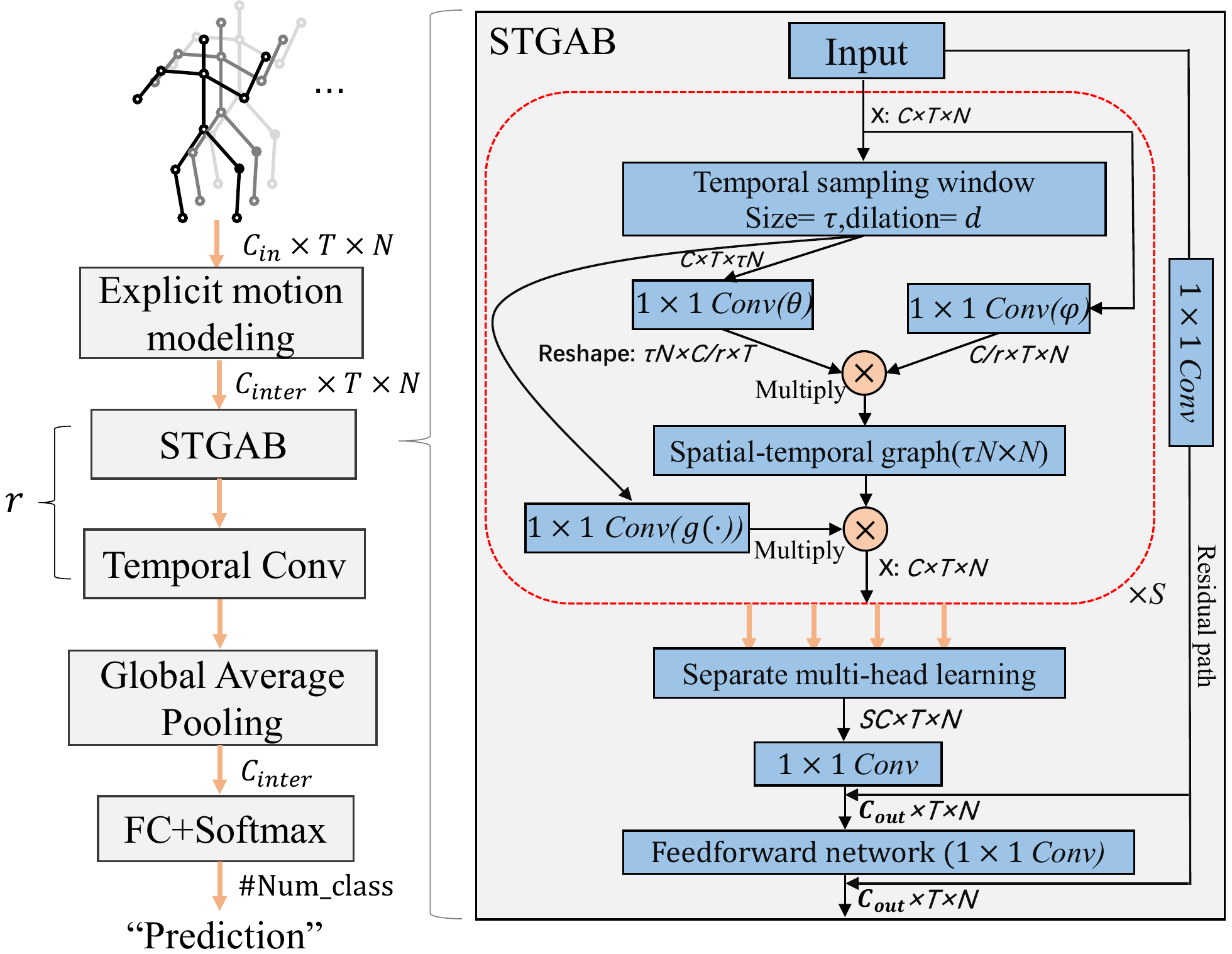} % Reduce the figure size so that it is slightly narrower than the column. Don't use precise values for figure width.This setup will avoid overfull boxes.
\caption{Architecture Overview of STGAT. After Explicit Motion Modeling, input data passes \textit{r} consecutive STGABs and temporal convolutions, followed by global average pooling, FC and softmax to perform classification. In each STGAB, \textit{S} heads are concatenated with two $1\times 1 $ convolutions to transform channels. For simplicity, BN and ReLU are neglected in the figure. Best viewed with color.}
\label{fig5}
\end{figure}
\subsection{Model Architecture}
\subsubsection{Overall Architecture.}
Our method is a \textit{general} framework which can be flexibly constructed upon self-attention operators or previous attention-based approaches. To demonstrate the effectiveness of STGAT, we choose DSTA-Net~\cite{shi2020decoupled} as our baseline, which employs fully attention-based spatial-only modules and temporal-only modules alternatively. It's already among the top-performing methods currently. As illustrated in fig.\ref{fig5}, our model is consisted of \textit{r} spatial-temporal graph attention blocks (STGABs) followed by a series of global average pooling, fully connected and softmax operations to perform prediction. Each STGAB contains a spatial-temporal graph attention module to capture regional cross-spacetime dependencies and a temporal convolution module to perform long-range temporal information aggregation. For simplicity, BN and ReLU and neglected in the figure.
%\subsubsection{Separate multi-head learning.}
%Our Separate Learning strategy can be combined with multi-head mechanism flexibly. We set one head as a single group. For each group, it will independently weight edges owning the same distance while the distance set for each group is different. 
\subsubsection{Multiple-Stream Fusion.}
Inspired by 2s-AGCN~\cite{shi2019two}, beneficial bone information is modeled by another stream. Following our baseline, we also calculate motion for both joint and bone streams to obtain another two motion streams. The softmax scores of input streams are averaged to give the final prediction.
  
\section{Experiments}
\subsection{Datasets}
\textbf{NTU RGB+D 60 and NTU RGB+D 120.} NTU RGB+D 60~\cite{shahroudy2016ntu} and NTU RGB+D 120~\cite{liu2019ntu} are both large-scale widely used action recognition datasets which separately contain 56578 action sequences over 60 classes and 113945 action sequences from 120 classes. All samples are performed by 40 or 106 different volunteers modeled by their 3D locations with \textit{N}=25. Two experiment setups are recommended by the author: (1) Cross-subject(X-Sub), where samples are split into training and testing groups with respect to subjects. (2) Cross-view(X-View) for NTU RGB+D 60 or Cross-Setup (X-Set) for NTU RGB+D 120, where action sequences are divided by their shooting cameras or camera setups.
  
\textbf{Kinetics Skeleton 400.} The Kinetics Skeleton 400 Dataset~\cite{kay2017kinetics} is sampled from the Kinetics-400 dataset by Openpose pose estimation toolkit~\cite{cao2017realtime}. It contains 240,436 training and 19,796 testing sequences. Every action sequence contains 18 joints wth their 2D locations and 1D confidence score at each time step.
\subsection{Training Details}
\label{sec:Training Details}
We use 8 STGABs in our experiments with their channels in \{64, 64, 128, 128, 256, 256, 256 and 256\} respectively. We randomly/uniformly sample 150 frames and then randomly/centrally crop 128 frames for training/test splits. The learning rate is 0.1 and divided by 10 at epochs \{60, 90\}, \{60, 90\}, and \{80, 110\} for NTU RGB+D Datasets, and Kinetics Skeleton 400 with total 120/180 epochs. The temporal convolution following STGAB owns kernel size 7. We combine our Separate Learning Strategy with multi-head mechanism and set one head as a single group. Our model is trained on a two-2080ti machine with batch size 32 and weight decay 0.0005 in PyTorch framework. Note that, no data augmentation technique is adopted for fair comparison. 

\subsection{Ablation Study}

\begin{minipage}{\textwidth}
\begin{minipage}[t]{0.53\textwidth}
\makeatletter\def\@captype{table}
\centering
          \begin{tabular}[t]{l|c} 
          \hline
          \textbf{Model configurations}  & \textbf{Top-1} \\
          \hline
          \hline
          Baseline  & 88.2\\
          STGAT & \textbf{88.0}\\
          \hline
          \hline
          w/ SL  & 88.6\\
          w/ SL + EMM & 89.4 \\
          w/ SL + EMM + $ T_{DTW}$  & \textbf{90.2}\\
          w/ SL + EMM + ATW & 90.1\\
          w/ SL + EMM + $T_{DTW}$ + ATW & 90.1\\
          Baseline w/ SL + EMM + $T_{DTW}$ & 89.0\\
          \hline
          Spatial connections (Baseline) & 88.2 \\
          Dense connections within $\tau$ frames & 87.2 \\
          Repeating $\tau$ times along time & 89.0\\
          STGAT & \textbf{90.2} \\
          \hline
        \end{tabular}
        %\end{subtable}    
        \caption{Ablations on NTU RGB+D 60 dataset. Top-1 accuracy(\%) is reported. We first compare STGAT with our baseline and then equip it with proposed strategies step by step to verify their effectiveness. Here, SL denotes Separate Learning strategy and EMM means Explicit Motion Modeling. $T_{DTW}$ represents the learnable parameter to structurally weight edges while ATW denotes Adaptive Temporal Weighting.}
        \label{table1} 
        \end{minipage}
        \begin{minipage}[t]{0.46\textwidth}
        \makeatletter\def\@captype{table}
        \centering
        \begin{tabular}[t]{l|c} 
        \hline
        \textbf{Model configurations}  & \textbf{Top-1} \\
          \hline
          \hline
          %STGAT$^{\dag}$ & \\
          \textit{S} = 1 & 89.0\\
          \textit{S} = 4 & 89.7\\  
          \textit{S} = 8 & \textbf{90.2}\\ 
          \textit{S} = 16 & 89.6\\  
          %\textbf{Window size $\tau$} & \textbf{Dilation factor \textit{d}} & \textbf{Top-1} \\
          \hline
          $\tau$=1      &   89.0  \\
          $\tau$=3, \textit{d}=1 & \textbf{90.2}\\ 
          $\tau$=3, \textit{d}=2 & 89.6   \\ 
          $\tau$=5, \textit{d}=1 & 89.2\\ 
          $\tau$=5, \textit{d}=2 & 89.1\\
          $\tau$=7, \textit{d}=1 & 89.1\\
          2 Paths, $\tau$=\{3,5\}, \textit{d}=\{1,1\} & 89.3\\
          2 Paths, $\tau$=\{3,5\}, \textit{d}=\{1,2\} & 89.1\\
          \hline    
          \end{tabular}
          %\end{subtable}    
          \caption{Ablations on NTU RGB+D 60 dataset. Top-1 accuracy(\%) is reported. We compare various STGAT settings, including different values of \textit{S}, $\tau$, \textit{d}. STGAT is configured with the best setting tested previously, i.e. STGAT w/ SL + EMM + $T_{DTW}$. }
          \label{table2} 
    \end{minipage}
    \end{minipage}  
\subsubsection{Effectiveness of proposed strategies.}

We incrementally build up the model with its individual components to verify their effectiveness and list the results in tab.~\ref{table1}. Here, SL denotes Separate Learning strategy and EMM means Explicit Motion Modeling. $T_{DTW}$ represents the learnable parameter to \textit{structurally} weight edges while ATW denotes Adaptive Temporal Weighting. Compared to our baseline, directly building spatial-temporal graphs with STGAT causes 0.2\% performance decrease. This demonstrates local feature redundancy severely hurts modeling short-term motion. We then testify the effectiveness of our proposed strategies for reducing local redundancy. Equipped with SL, STGAT obtains 0.6\% performance boost. Through SL, STGAT deals with small groups of joints and is easier to distinguish vital connections. Combined with EMM, STGAT achieves 0.8\% accuracy boost, which demonstrates EMM makes effect by capturing subtle motion instead of raw spare features. Next, we test the effect of Dynamic Temporal Weighting strategy. STGAT gains 0.8\% performance boost from $T_{DTW}$ which \textit{structurally} weights frames along temporal dimension. ATW also performs well and brings 0.7\% performance boost. These results indicate our strategy corrects the equal weighting problem. We also try to combine $T_{DTW}$ with ATW to perform structural and adaptive weighting together, but don't witness further promotion. We speculate that they may overemphasize the weights. $T_{DTW}$ is adopted as default due to its high performance and low computation cost. In the following, we make a comparative test to verify the effectiveness of STGAT. We don't build spatial-temporal graphs but directly equip our baseline with proposed strategies. In this condition, SL and EMM will still make effect. As shown in tab.~\ref{table1}, it achieves 89.0\% (+0.8\%) but performs lower than STGAT (90.2\%, +2.0\%), which demonstrates STGAT can better capture local spatial-temporal dependencies for its local spatial-temporal modeling. Finally, we examine the effect of spatial-temporal modeling for STGAT. We compare it with spatial connections (baseline), dense connections among all joints in $\tau$ frames, and simply repeating spatial graph $\tau$ times along temporal dimension. Our STGAT outperforms all of them by a large margin ($\geq$ 1.2\%) which demonstrates the effectiveness of adaptive computation for edges in a local spatial-temporal neighborhood. 
\subsubsection{Study on model structure.}

We then test the choices of some hyperparameters in STGAT and give their results in tab.~\ref{table2}. We offer \textit{S}, the number of heads, with four choices in \{1, 4, 8 and 16\}. We observe that larger \textit{S} gives higher performance which indicates the effects of multiple built graphs, and the performance reaches top when \textit{S} equals 8. For other STGAT configurations, like $\tau$ and \textit{d}, we test different combinations of them. $\tau$=1 means no temporal information is incorporated and only spatial modeling is performed. We first observe STGAT outperforms baseline when $\tau$=1 and benefits from larger $\tau$, which demonstrates the effect of local spatial-temporal dependencies. The gain of $\tau$ starts to decrease when it equals 5. We infer large $\tau$ makes similar effect with temporal-only modules and causes redundancy. Considering \textit{d}, we observe accuracy drop when \textit{d} increases, which shows closer frames are much more relevant in a local spatial-temporal neighborhood. We further construct two STGAT paths to make comparison with MS-G3D~\cite{liu2020disentangling}. MS-G3D obtains performance boost by incorporating more paths for multiscale temporal information. But STGAT already earns better performance through one path with fewer computations and parameters than MS-G3D for its superior spatial-temporal modeling ability.

\subsubsection{Effect on classifying similar actions.}

We demonstrate the effect of STGAT on classifying similar patterns in tab.~\ref{table3}. We compare STGAT with baseline and another spatial-temporal modeling method, MS-G3D, upon several similar action pairs from NTU RGB+D 60 Dataset. Compared with MS-G3D, STGAT outperforms it in all action pairs with up to 9.2\% performance boost. MS-G3D fails to distinguish short-term motion with equal weights for neighboring joints, which introduces too much redundancy, even with two paths. Compared with baseline, STGAT obtains notable improvements, up to 12.5\%, upon similar action pairs. These results verify the effectiveness of our STGAT on classifying similar action pairs owing to its adaptive spatial-temporal modeling. Please refer to the supplementary material for more examples and the full table.
\begin{table}
\centering
\vspace{-5px}
  \begin{tabular}{c|c|c|c}
  \hline
  Classes & MS-G3D    & Baseline & STGAT \\
  \hline
  Average of all classes & 89.4  & 88.2 & 90.2 \\ 
  \hline
  eat meal/snack  &    67.6   & 70.9   & 74.9 (\textbf{+7.3,+4.0})  \\
  brushing teeth  &   87.5    & 82.1   & 89.0 (\textbf{+1.5,+6.9})  \\
  \hline
  reading   &   68.5  & 60.1   & 71.6 (\textbf{+3.1,+11.5}) \\
  writing  &   54.4   & 51.1   & 63.6 (\textbf{+9.2,+12.5})  \\
  \hline
  taking a selfie &   89.1   & 86.6  & 90.5 (\textbf{+1.4,+3.9}) \\
  salute   &   89.9  & 88.0   & 93.1 (\textbf{+3.2,+5.1})  \\
  \hline
  \end{tabular}  
  \caption{Class-wise Accuracy(\%) Comparison for MS-G3D~\cite{liu2020disentangling}, baseline and STGAT on similar action pairs from NTU RGB+D 60 dataset with X-Sub Setting.}
  \vspace{-25px}
  \label{table3}  
  \end{table}
  
\subsubsection{Comparison with previous methods with short-term modeling.}
\label{sec:sec4.3}
We compare STGAT with MS-GCN~\cite{liu2020disentangling} which assigns equal weights for local neighbors for spatial-temporal modeling. ~\cite{liu2020disentangling} also proposes MSTCN consisted of temporal convolutions with different dilations for multiscale temporal information. As shown in tab.~\ref{table4}, STGAT outperforms baseline as well as MSGCN by a large margin (+2.0\%, +2.4\%). When equipped with MSTCN, either baseline or MSGCN obtains satisfactory performance boost (+0.4\%, +1.2\%), accompanied by notable increase in parameters and GFLOPs. STGAT outperforms MSTCN hybrids by at least 1.2\% with fewer parameters and GFLOPs and similar memory usage, which demonstrates the efficacy and efficiency of STGAT for capturing both short-term and long-term dependencies.
\begin{table}
    \centering
    \vspace{-5px}
    \begin{tabular}{cccccc}
    \hline
    Methods & \#Params(M) & GFLOPs & Training Hours & Memory(G) & Top-1(\%) \\ 
    \hline
    Baseline & 2.4  & 10.2 & 6.0 & 16.0 & 88.2\\
    MS-GCN~\cite{liu2020disentangling}  & 1.4 & 12.4 & 8.6 & 18.8& 87.8 \\
    \hline
    Baseline+3*MSTCN~\cite{liu2020disentangling} & 3.7  & 15.9 & 7.1 & 23.5 & 88.6\\
    MS-GCN+3*MSTCN~\cite{liu2020disentangling} & 2.7 & 18.1 & 12.2& 25.4& 89.0\\
    \hline
    STGAT & 2.4 & 14.6 & 10.1 & 24.2 &  \textbf{90.2}\\
    \hline
    \end{tabular}  
    \caption{Comparison with previous methods with short-term modeling on NTU RGB+D 60 dataset with X-Sub Setting.}
    \label{table4}  
    \vspace{-25px}
    \end{table}

\subsubsection{Visualizations.} 
\begin{figure}[t]
    \centering
    \includegraphics[width=\textwidth]{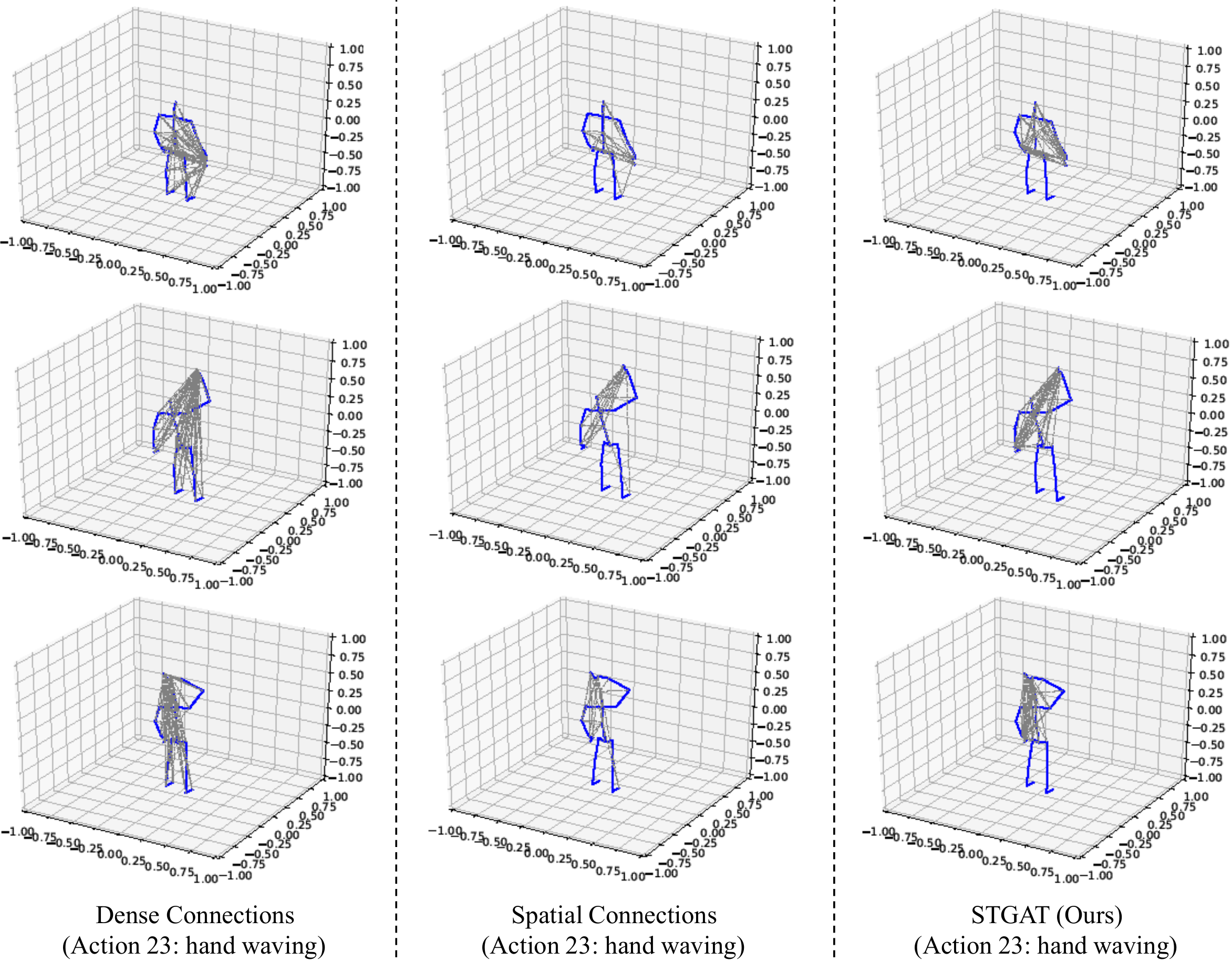} % Reduce the figure size so that it is slightly narrower than the column. Don't use precise values for figure width.This setup will avoid overfull boxes.
    \caption{Visualizations for generated edges by STGAT, as well as dense connections among all joints, and spatial connections from adaptively computed spatial graph.}
    \label{fig8}
    \end{figure}

We visualize the generated edges by STGAT, dense connections among all joints, and spatial connections (Baseline), in fig.~\ref{fig8} to give more analysis on the effect of STGAT. We select the action "hand waving" as an example. The left hand is the most important joint of action "hand waving" which forms the movement patterns of this action. Edges generated by STGAT all lie in the upper body and especially focuses on the left hand. In contrast, the spatial connections (Baseline) are more sparse and some land on the bottom body which has no movements. Dense connections have no focus and put same attention on all joints.

  \begin{table}[t]   
      \centering
      \begin{tabular}{ccc} 
      \hline
      \multicolumn{3}{c}{\textbf{NTU RGB+D 60}} \\
      Methods & X-Sub & X-View  \\
      \hline
      %HCN~\cite{li2018co}  & 86.5 & 91.1\\
      AS-GCN~\cite{li2019actional} & 86.8 & 94.2\\
      2s-AGCN~\cite{shi2019two} &  88.5 & 95.1\\
      AGC-LSTM~\cite{si2019attention} &  89.2 & 95.0\\
      GCN-NAS~\cite{peng2020learning} & 89.4 & 95.7 \\
      DGNN~\cite{shi2019skeleton1} &  89.9 & 96.1\\
      MS-AAGCN~\cite{shi2019skeleton} & 90.0 & 96.2\\
      Shift-GCN~\cite{cheng2020skeleton} & 90.7 & 96.5 \\
      %DC-GCN~\cite{chengdecoupling} & 90.8 & 96.6\\
      PA-ResGCN~\cite{song2020stronger} & 90.9 & 96.0 \\
      Dynamic GCN~\cite{ye2020dynamic} & 91.5 & 96.0 \\
      MS-G3D Net~\cite{liu2020disentangling}& 91.5 & 96.2 \\
      4s DSTA-Net~\cite{shi2020decoupled} & 91.5 & 96.4 \\
      4s MST-GCN~\cite{chen2021multi} &91.5 & 96.6 \\
      GCsT~\cite{bai2021gcst}  &  91.6 & 96.2 \\
      \hline
      STGAT(Joint) & 90.2 &  95.4 \\
      STGAT(Bone) & 90.6 &  95.8\\
      \textbf{2s-STGAT} & \textbf{92.2} &  \textbf{96.9} \\  
      \textbf{4s-STGAT} & \textbf{92.8} &  \textbf{97.3} \\   
      \hline
      \end{tabular}    
      \caption{Comparison of STGAT against other state-of-the-art methods on the NTU RGB+D 60 dataset. Top-1 accuracy(\%) is reported.} 
      \label{table5}
      \vspace{-25px}
      \end{table}
 
\subsection{Comparison with the state-of-the-art}
Table~\ref{table5}, ~\ref{table6} and ~\ref{table7} compare our final model with other state-of-the-art methods on three large-scale datasets NTU RGB+D 60~\cite{shahroudy2016ntu}, NTU RGB+D 120~\cite{shahroudy2016ntu} and the Kinetics Skeleton 400~\cite{kay2017kinetics}. These methods include CNN-based methods~\cite{li2018co}, RNN-based methods~\cite{liu2016spatio,si2019attention}, graph networks with predefined graph structure~\cite{li2019actional,liu2020disentangling,shi2019skeleton1,yan2018spatial,cheng2020skeleton} and graph networks with adaptive adjacent matrix~\cite{shi2019two,shi2019skeleton,shi2020decoupled,zhang2020semantics,ye2020dynamic}. We first observe that most of graph-based methods gain better performance than non-graph methods by better modeling relationships. We then see graph methods with adaptive graph structure~\cite{shi2019two,shi2019skeleton,shi2020decoupled,zhang2020semantics,ye2020dynamic} or multi-scale feature aggregation~\cite{liu2020disentangling} consistently perform better than previous methods, indicating the necessity to capture long-range dependencies. Our STGAT makes a step further to capture cross-spacetime information and outperforms all existing methods especially when MS-G3D aggregates spatial-temporal features with two paths. 
\section{Conclusion}
\vspace{-5px}
In this paper, we propose a general framework, named STGAT, to model cross-spacetime information. To reduce local spatial-temporal feature redundancy, we propose three simple modules on narrowing the scope of self-attention mechanism, dynamically weighting frames and separating subtle motion from static features. STGAT exceeds previous state-of-the-arts on three large-scale datasets and makes further progress over classifying similar action pairs, witnessed by both qualitative and quantitative results.
\vspace{10px}

\begin{minipage}{\textwidth}
\begin{minipage}[t]{0.46\textwidth}
\makeatletter\def\@captype{table}
\centering
\begin{tabular}{ccc} 
    \hline
    \multicolumn{3}{c}{\textbf{NTU RGB+D 120}}\\
    Methods & X-Sub & X-Set\\
    \hline
    %ST-LSTM~\cite{liu2016spatio} &  55.7 & 57.9\\
    SGN~\cite{zhang2020semantics}	& 79.2	& 81.5\\
    2s-AGCN~\cite{shi2019two} &  82.9 & 84.9 \\
    Shift-GCN~\cite{cheng2020skeleton} & 85.9 & 87.6 \\
    4s DSTA-Net~\cite{shi2020decoupled} & 86.6 & 89.0 \\
    MS-G3D Net~\cite{liu2020disentangling} &  86.9 & 88.4\\
    PA-ResGCN~\cite{song2020stronger} & 87.3 & 88.3 \\
    Dynamic GCN~\cite{ye2020dynamic} &   87.3  &  88.6\\
    4s MST-GCN~\cite{chen2021multi} &87.5& 88.8 \\
    GCsT~\cite{bai2021gcst} & 87.7 & 89.3 \\
    \hline
    \textbf{4s-STGAT} & \textbf{88.7} &  \textbf{90.4} \\
    \hline
    \end{tabular}    
    \caption{Comparison of STGAT against other state-of-the-art methods on the NTU RGB+D 120 dataset. Top-1 accuracy(\%) is reported.} 
    \label{table6}
\end{minipage}
\begin{minipage}[t]{0.46\textwidth}
\makeatletter\def\@captype{table}
\centering
\begin{tabular}{ccc} 
    \hline
    \multicolumn{3}{c}{\textbf{Kinetics Skeleton 400}} \\
    Methods & Top-1 & Top-5 \\
    \hline
    ST-GCN~\cite{yan2018spatial} &  30.7 & 52.8\\
    %AS-GCN~\cite{li2019actional} &  34.8 & 56.5\\
    2s-AGCN~\cite{shi2019two} &  36.1 & 58.7\\
    DGNN~\cite{shi2019skeleton1} &  36.9 & 59.6\\
    GCN-NAS~\cite{peng2020learning} & 37.1 & 60.1 \\
    MS-AAGCN~\cite{shi2019skeleton} & 37.8 & 61.0\\
    Dynamic GCN~\cite{ye2020dynamic} &  37.9 & 61.3\\
    MS-G3D Net~\cite{liu2020disentangling} & 38.0 & 60.9\\
    4s MST-GCN~\cite{chen2021multi} &38.1& 60.8 \\
    \hline
    \textbf{4s-STGAT} & \textbf{39.2} & \textbf{62.8}  \\
    \hline
    \end{tabular}    
    \caption{Comparison of STGAT against other state-of-the-art methods on the Kinetics Skeleton 400 dataset. Top-1 and Top-5 accuracies(\%) are reported.} 
    \label{table7}
\end{minipage}
\end{minipage}
  
\bibliographystyle{splncs04}
\bibliography{ref}
\appendix

\title{Supplementary Material for Spatial Temporal Graph Attention Network for Skeleton-Based Action Recognition} % Replace with your title

\titlerunning{STGAT for Skeleton-based Action Recognition}

\author{Lianyu Hu\inst{1}\orcidlink{0000-0003-2470-8110} \and
Shenglan Liu\inst{2}\orcidlink{0000-0003-3823-4200} \and
Wei Feng\inst{1}\orcidlink{0000-0003-3809-1086}}
\authorrunning{Lianyu et al.}
\institute{Tianjin University \and Dalian University of Technology
%\email{hly2021@tju.edu.cn;liusl@dlut.edu.cn;wfeng@ieee.org}
}
\maketitle

\section{Implementations of Self-attention modules}
Under the formulation defined in Eq.4 in our manuscript, we give four different flexible instantiations for the pairwise function \textit{f}. Other alternative choices are also applicable and may give better results. We employ Embedded Gaussian function as default for \textit{f}.

\textbf{Gaussian.} A natural choice for \textit{f} is the widely used Gaussian function which can be expressed as:
\begin{equation}
    \label{e6}
    f(x_{i},x_{j})= e^{x_{i}^{T}x_{j}}.
\end{equation}
Here $x_{i}^{T}x_{j}$ is the dot-product affinity of two position inputs for simplicity. Other distance measurements are also applicable. Correspondingly, the normalization factor $C(x)$ is set as $\sum_{\forall j} f(x_{i},x_{j})$.

\textbf{Embedded Gaussian.} A simple extension of the Gaussian function for \textit{f} is to compute it in an embedding space, as:
\begin{equation}
    \label{e7}
    f(x_{i},x_{j})= e^{\theta(x_{i})^{T}\phi(x_{j})}
\end{equation}
where $\theta(x)$ and $\phi(x)$ are two embeddings. In practice, they are always implemented with 1$\times$1 convolutions. $C(x)$ is set as $\sum_{\forall j} f(x_{i},x_{j})$. A reason for the popularity of the Embedded Gaussian function is that it can be easily implemented with the softmax function, which both serves as the activation function and adds normalization. As we can see, for a given position \textit{i}, once we have set $C(x)$ properly, $\frac{1}{C(x)}e^{\theta(x_{i})^{T}\phi(x_{j})}$ is exactly the form of the softmax function along dimension \textit{j}.

\textbf{Dot Product.} Except for the series of Gaussian functions, a dot-product affinity can also be adopted, as:
\begin{equation}
    \label{e8}
    f(x_{i},x_{j})= \theta(x_{i})^{T}\phi(x_{j}).
\end{equation}
In this case, $C(x)$ is set as the number of possible positions in \textit{x} for convenience in gradient computation. Other activation functions can be followed.

\textbf{Concatenation.} Concatenation function is always implemented as:
\begin{equation}
    \label{e9}
    f(x_{i},x_{j})= \sigma({W_f[\theta(x_{i}),\phi(x_{j})]})
\end{equation}
where $[\cdot,\cdot]$ represents concatenation operation and $W_f$ is a weight vector projecting their concatenated result into a scalar. Besides, $\sigma$ serves as an activation function that is usually ReLU or LeakyReLU in practice.

\begin{figure}[t]
    \centering
    \includegraphics[width=0.7\columnwidth]{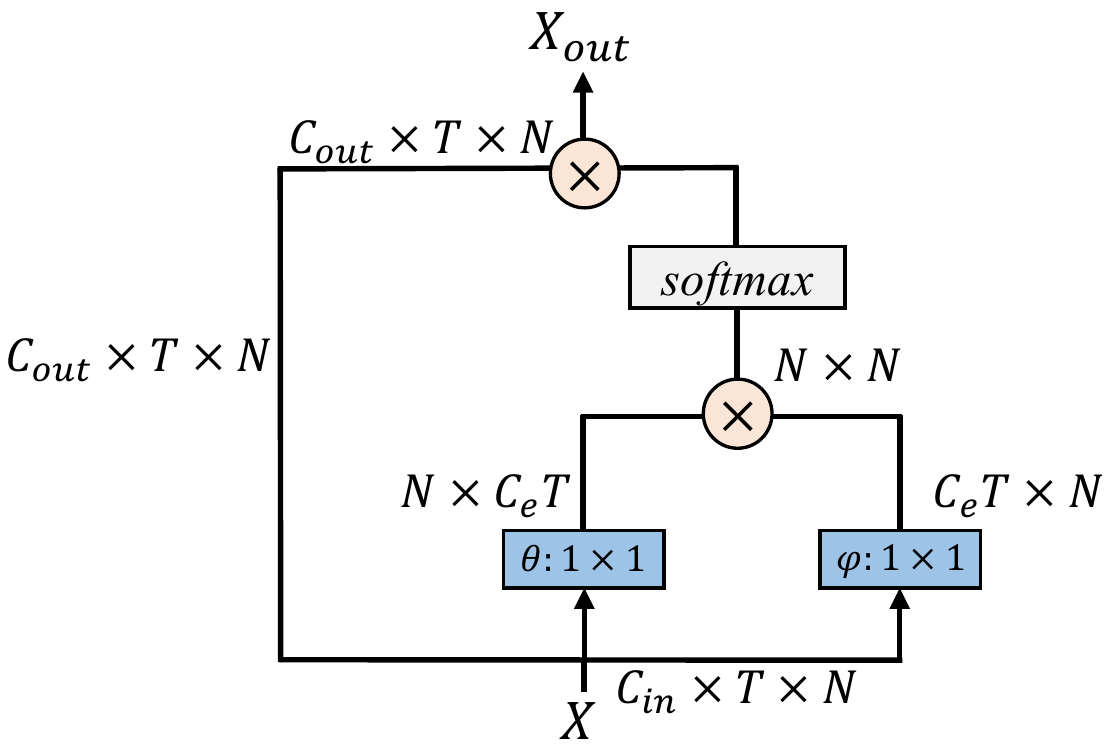} % Reduce the figure size so that it is slightly narrower than the column. Don't use precise values for figure width.This setup will avoid overfull boxes.
    \caption{A self-attention block with Embedded Gaussian version. $\theta$ and $\phi$ are implemented with 1$\times$1 convolutions. $g(\cdot)=1$. $C_e$ is adopted to reduce computation costs. The vanilla Gaussian version can be obtained by removing $\theta$ and $\phi$, and the Dot-product version can be done by replacing softmax with a scaling factor 1/N. }.
    \label{fig2}
\end{figure}

Under the formulation defined above, a self-attention module can be implemented as shown in fig.~\ref{fig2} in practice. Here, $\theta(\cdot)$ and $\phi(\cdot)$ are instantiated as 1$\times$1 convolutions and $C_e$ is adopted to reduce computation costs. $g(\cdot)=1$. Empirically, $C_e$ is defined as $C_{out}/r$ where \textit{r} is an adjustable factor to reduce computations.

\section{Implementation for Adaptive Temporal Weighting Module}
\begin{figure}[h]
    \centering
    \includegraphics[width=0.7\columnwidth]{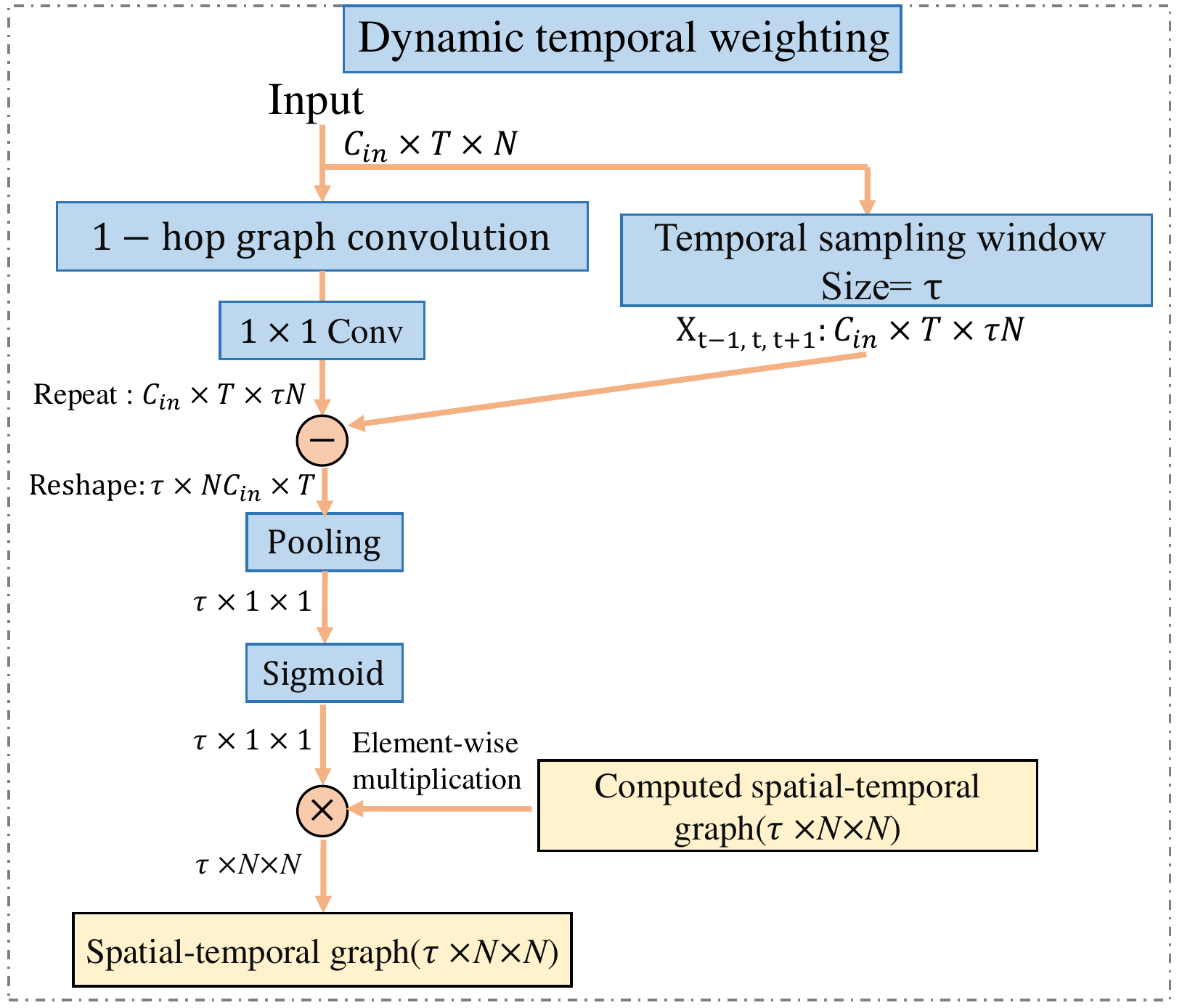} % Reduce the figure size so that it is slightly narrower than the column. Don't use precise values for figure width.This setup will avoid overfull boxes.
    \caption{ Illustration of our proposed adaptively Dynamic Temporal Weighting strategy. The goal is to weight frames in spatial-temporal graph with an adaptively computed factor with size $\tau\times 1\times 1 $.}
    \label{fig3}
\end{figure}

We give the implementation for the adaptive temporal weighting module in the fig.~\ref{fig3}. We first perform 1-hop graph convolutions for current frame $X^{t}$ to aggregate local spatial features. Then, a $1\times 1$ convolution kernel is utilized to enhance features of current frame $X^{t}$. We subtract sampled features of regional neighbors from those of current frame to obtain motion features. Finally, after pooling along channel and temporal dimension, they are passed through a sigmoid function to weight frames in local spatial-temporal region.

\section{Class-wise Accuracy Comparison}
Table.~\ref{table2} offers class-wise accuracy comparison for MS-G3D~\cite{liu2020disentangling}, our baseline and STGAT on the Cross Subject setting of the NTU RGB+D 60~\cite{shahroudy2016ntu} dataset. Specifically, accuracy decrease is marked in red while accuracy increase is marked with green. It can first be observed that MS-G3D and baseline have achieved great performance on most of the classes with 89.4\% and 88.2\% average, respectively. Our final model consistently outperforms MS-G3D and baseline in 38/60 and 46/60 classes. Especially, on several similar action pairs like 'eat meal/snack' against 'brushing teeth', 'reading' against 'writing' and 'taking a selfie' against 'salute', STGAT makes great improvements (+7.3\%/+1.5\%, +3.1\%/+9.2\%, +1.4\%/+3.2\%) over MS-G3D and (+4.0\%/+6.9\%, +11.5\%/+12.5\%, +3.9\%/+5.1\%) over baseline due to the special care of short-term dependencies.

\clearpage
\begin{centering}
    \begin{longtable}{c|c|ccc}
        \hline
        Classes                           & \#ID & MS-G3D & Baseline                        & STGAT                                                     \\
        \hline
        Average of all classes            & -    & 89.4  & \makecell[c]{88.2}              & \makecell[c]{90.2}                                        \\
        \hline
        drink water                       & 0    & 85.8  & 85.4   & 87.6 (\textcolor{green}{+1.8}, \textcolor{green}{+2.2})  \\
        eat meal/snack                    & 1    & 67.6  & 70.9   & 74.9 (\textcolor{green}{+7.3}, \textcolor{green}{+4.0})  \\
        brushing teeth                    & 2    & 87.5  & 82.1     & 89.0 (\textcolor{green}{+1.5}, \textcolor{green}{+6.9})   \\
        brushing hair                     & 3    & 90.8  & 88.3  & 92.7 (\textcolor{green}{+1.9}, \textcolor{green}{+4.4})  \\
        drop                              & 4    & 89.8 & 91.3  & 93.5 (\textcolor{green}{+3.7}, \textcolor{green}{+2.2})  \\
        pickup                            & 5    & 91.6  & 92.4     & 92.7 (\textcolor{green}{+1.1}, \textcolor{green}{+0.3})   \\
        throw                             & 6    & 94.2  & 92.7  & 88.4 (\textcolor{red}{-5.8}, \textcolor{red}{-4.3})    \\
        sitting down                      & 7    & 94.5  & 92.7   & 97.4 (\textcolor{green}{+2.9}, \textcolor{green}{+4.7})   \\
        standing up                       & 8    & 98.2  & 97.1   & 97.4 (\textcolor{red}{-0.8}, \textcolor{green}{+0.3})   \\
        clapping                          & 9    & 82.8  & 78.4   & 81 (\textcolor{red}{-1.8}, \textcolor{green}{+2.6})    \\
        reading                           & 10   & 68.5  & 60.1   & 71.6 (\textcolor{green}{+3.1}, \textcolor{green}{+13.1})   \\
        writing                           & 11   & 54.4  & 51.1  & 63.6 (\textcolor{green}{+9.2}, \textcolor{green}{+12.5}) \\
        tear up paper                     & 12   & 90.4  & 91.1  & 86.3 (\textcolor{red}{-4.1}, \textcolor{red}{-4.8})     \\
        wear jacket                       & 13   & 98.9  & 96.4   & 98.5 (\textcolor{red}{-0.4}, \textcolor{green}{+2.1})   \\
        take off jacket                   & 14   & 96.7  & 96.4   & 97.1 (\textcolor{green}{+0.4}, \textcolor{green}{+0.7})   \\
        wear a shoe                       & 15   & 87.2  & 85.0  & 84.2 (\textcolor{red}{-3.0}, \textcolor{red}{-0.8})    \\
        take off a shoe                   & 16   & 74.5  & 82.8   & 78.5 (\textcolor{green}{+4.0}, \textcolor{red}{-4.3})     \\
        wear on glasses                   & 17   & 91.6  & 91.9  & 89.4 (\textcolor{red}{-2.2}, \textcolor{red}{-2.5})     \\
        take off glasses                  & 18   & 91.6 & 93.8  & 90.5 (\textcolor{red}{-1.1}, \textcolor{red}{-3.3})    \\
        put on a hat/cap                  & 19   & 95.2  & 94.1   & 94.1 (\textcolor{red}{-1.1}, \textcolor{green}{+0.0})   \\
        take off a hat/cap                & 20   & 94.9  & 94.1   & 97.1 (\textcolor{green}{+2.2}, \textcolor{green}{+3.0})   \\
        cheer up                          & 21   & 95.3 & 91.6   & 92.0 (\textcolor{red}{-2.3}, \textcolor{green}{+0.4})   \\
        hand waving                       & 22   & 92.3  & 89.4 ( & 93.0 (\textcolor{green}{+0.7}, \textcolor{green}{+3.6})  \\
        kicking something                 & 23   & 96.4  & 95.3   & 97.1 (\textcolor{green}{+0.7}, \textcolor{green}{+1.8})  \\
        reach into pocket                 & 24   & 83.9  & 81.8   & 80.3 (\textcolor{red}{-3.6}, \textcolor{red}{-1.5})     \\
        hopping (one foot jumping)        & 25   & 97.5  & 96.4   & 98.9 (\textcolor{green}{+1.4}, \textcolor{green}{+2.5})  \\
        jump up                           & 26   & 99.3  & 97.8   & 100 (\textcolor{green}{+0.7}, \textcolor{green}{+2.2})    \\
        make a phone call/answer phone    & 27   & 86.2  & 82.5   & 83.6 (\textcolor{red}{-2.6}, \textcolor{green}{+1.1})  \\
        playing with phone/tablet         & 28   & 72.4  & 68.4     & 82.5  (\textcolor{green}{+10.1}, \textcolor{green}{+14.1}) \\
        typing on a keyboard              & 29   & 67.3  & 75.6  & 69.8 (\textcolor{green}{+2.5}, \textcolor{red}{-5.8})     \\
        pointing to something with finger & 30   & 81.2  & 75.0     & 80.8 (\textcolor{red}{-0.4}, \textcolor{green}{+5.8})   \\
        taking a selfie                   & 31   & 89.1  & 86.6  & 90.5 (\textcolor{green}{+1.4}, \textcolor{green}{+3.9})  \\
        check time (from watch)           & 32   & 88.4  & 84.4   & 91.7 (\textcolor{green}{+3.3}, \textcolor{green}{+7.3})   \\
        rub two hands together            & 33   & 86.6  & 88.8  & 88.4 (\textcolor{green}{+1.8}, \textcolor{red}{-0.4})    \\
        nod head/bow                      & 34   & 94.6  & 93.1   & 96.0 (\textcolor{green}{+1.4}, \textcolor{green}{+2.9})   \\
        shake head                        & 35   & 94.5  & 93.8  & 93.8 (\textcolor{red}{-0.7}, \textcolor{green}{+0.0})  \\
        wipe face                         & 36   & 87.3  & 81.5   & 88.4  (\textcolor{green}{+0.9}, \textcolor{green}{+6.9})  \\
        salute                            & 37   & 89.9 & 88.0   & 93.1 (\textcolor{green}{+3.2}, \textcolor{green}{+5.1})  \\
        put the palms together            & 38   & 92.4  & 92.8   & 93.1 (\textcolor{green}{+0.7}, \textcolor{green}{+0.3})   \\
        cross hands in front (say stop)   & 39   & 94.9  & 89.5   & 94.6 (\textcolor{red}{-0.3}, \textcolor{green}{+5.1})  \\
        sneeze/cough                      & 40   & 75.7  & 73.2  & 75.0 (\textcolor{red}{-0.7}, \textcolor{green}{+1.8})  \\
        staggering                        & 41   & 99.6  & 96.4   & 97.8 (\textcolor{red}{-1.8}, \textcolor{green}{+1.4})  \\
        falling                           & 42   & 97.8  & 95.6   & 97.8 (\textcolor{green}{+0.0}, \textcolor{green}{+2.2})  \\
        touch head (headache)             & 43   & 83.8  & 81.5  & 74.3 (\textcolor{red}{-9.5}, \textcolor{red}{-7.2})    \\
        touch chest                       & 44   & 90.2  & 90.2     & 91.3 (\textcolor{green}{+1.1}, \textcolor{green}{+1.1})   \\
        touch back (backache)             & 45   & 94.9  & 93.1   & 90.2 (\textcolor{red}{-4.7}, \textcolor{red}{-2.9})     \\
        touch neck (neckache)             & 46   & 86.2  & 80.4   & 86.5(\textcolor{green}{+0.3}, \textcolor{green}{+6.1})     \\
        nausea or vomiting condition      & 47   & 84.0  & 84.7   & 85.1 (\textcolor{green}{+1.1}, \textcolor{green}{+0.4})  \\
        use a fan /feeling warm           & 48   & 91.6  & 86.5   & 90.9 (\textcolor{red}{-0.7}, \textcolor{green}{+4.4})   \\
        punching/slapping other person    & 49   & 89.8  & 90.9  & 91.2 (\textcolor{green}{+1.4}, \textcolor{green}{+0.3})  \\
        kicking other person              & 50   & 92.0  & 92.0  & 94.9 (\textcolor{green}{+2.9}, \textcolor{green}{+2.9})  \\
        pushing other person              & 51   & 92.1  & 94.2 ( & 97.1 (\textcolor{green}{+5.0}, \textcolor{green}{+2.9})  \\
        pat on back of other person       & 52   & 92.4  & 80.1   & 96.7 (\textcolor{green}{+4.3}, \textcolor{green}{+16.6}) \\
        point finger at the other person  & 53   & 87.7  & 86.2   & 92.0 (\textcolor{green}{+4.3}, \textcolor{green}{+5.8})   \\
        hugging other person              & 54   & 98.5  & 96.4     & 98.5 (\textcolor{green}{+0.0}, \textcolor{green}{+2.1})   \\
        giving something to other person  & 55   & 93.1  & 91.7   & 93.8 (\textcolor{green}{+0.7}, \textcolor{green}{+2.1})  \\
        touch other person's pocket       & 56   & 92.0  & 89.8   & 93.8 (\textcolor{green}{+1.8}, \textcolor{green}{+4.0})   \\
        handshaking                       & 57   & 97.1  & 95.3     & 96.4 (\textcolor{red}{-0.7}, \textcolor{green}{+1.1})   \\
        walking towards each other        & 58   & 97.8  & 98.5   & 97.8 (\textcolor{green}{+0.0}, \textcolor{red}{-0.7})     \\
        walking apart from each other     & 59   & 96.0  & 95.7   & 94.6 (\textcolor{red}{-1.4}, \textcolor{red}{-1.1})     \\
        \hline
        %\captionsetup{justification=raggedright}
        \caption{Class-wise Accuracy Comparison for MS-G3D~\cite{liu2020disentangling}, our baseline (DSTA-Net~\cite{shi2020decoupled}) and STGAT on the Cross Subject setting of the NTU RGB+D 60 dataset, using the joint only. Values in the parentheses are accuracy differences compared to MS-G3D or baseline. Accuracy decrease is marked in red while accuracy increase is marked with green. STGAT outperforms MS-G3D and baseline in 38/60 and 46/60 action classes, respectively. Especially, on several similar action pairs like 'eat meal/snack' against 'brushing teeth', 'reading' against 'writing' and 'taking a selfie' against 'salute', STGAT makes great improvements.}
        \label{table2}%
    \end{longtable}%
\end{centering}

\end{document}